\documentclass[letterpaper, 10pt, conference, twoside]{style/ieeeconf}    

\makeatletter
\let\NAT@parse\undefined
\makeatother

\usepackage{dblfloatfix}

\usepackage[numbers,sectionbib,sort&compress]{natbib}
\usepackage{bm}
\usepackage{gensymb}
\usepackage{xcolor}
\usepackage{graphicx}
\usepackage{amsmath}
\usepackage{amssymb}
\usepackage{subcaption}
\usepackage{amsfonts}
\usepackage{siunitx}
\usepackage{booktabs}
\usepackage{makecell}
\usepackage{multirow}
\usepackage{upgreek}
\usepackage[font=small]{caption}
\usepackage[export]{adjustbox}
\usepackage{tikz}
\usepackage{tabularx}
\usepackage{svg}
\usepackage{sidecap} \sidecaptionvpos{figure}{c}
\usepackage{hyperref}
\hypersetup{colorlinks,breaklinks,
linkcolor=[rgb]{0.5,0.,0.},
citecolor=[rgb]{0.000,0.427,0.173},
urlcolor=[rgb]{0.031,0.318,0.612}}

\usepackage[nameinlink, capitalize]{cleveref}
\usepackage{color}
\definecolor{CommentPink}{rgb}{1,0.2,0.5}
\definecolor{CommentBlue}{rgb}{0,0,1}
\definecolor{CommentGreen}{rgb}{0,1,0}

\Crefname{section}{Sec.}{Sec.}
\Crefname{equation}{Eq.}{Eq.}

\usepackage[printonlyused,withpage,nolist,nohyperlinks]{acronym}
\begin{acronym}

\acro{2D}{two-dimensional}

\acro{3D}{three-dimensional}

\acro{AHRS}{attitude and heading reference system}

\acro{AUV}{autonomous underwater vehicle}

\acro{CPP}{Chinese Postman Problem}

\acro{DoF}{degree of freedom}
\acrodefplural{DoF}[DoFs]{degrees of freedom}

\acro{DVL}{Doppler velocity log}

\acro{FSM}{finite state machine}

\acro{IMU}{inertial measurement unit}

\acro{LBL}{Long Baseline}

\acro{MCM}{mine countermeasures}

\acro{MDP}{Markov decision process}
\acrodefplural{MDP}[MDPs]{Markov decision processes}

\acro{POMDP}{Partially Observable Markov Decision Process}
\acrodefplural{POMDP}[POMDPs]{Partially Observable Markov Decision Processes}

\acro{PRM}{Probabilistic Roadmap}
\acrodefplural{PRM}[PRM]{Probabilistic Roadmaps}

\acro{ROI}{region of interest}
\acrodefplural{ROI}[ROIs]{regions of interest}

\acro{ROS}{Robot Operating System}

\acro{ROV}{remotely operated vehicle}

\acro{RRT}{Rapidly-exploring Random Tree}
\acrodefplural{RRT}[RRTs]{Rapidly-exploring Random Trees}

\acro{SLAM}{Simultaneous Localisation and Mapping}

\acro{SSE}{sum of squared errors}

\acro{STOMP}{Stochastic Trajectory Optimization for Motion Planning}

\acro{TRN}{Terrain-Relative Navigation}

\acro{UAV}{unmanned aerial vehicle}

\acro{USBL}{Ultra-Short Baseline}

\acro{IPP}{informative path planning}

\acro{FoV}{field of view}
\acrodefplural{FoV}[FoVs]{fields of view}

\acro{CDF}{cumulative distribution function}

\acro{ML}{maximum likelihood}

\acro{RMSE}{Root Mean Squared Error}
\acro{MLL}{Mean Log Loss}

\acro{GP}{Gaussian Process}
\acrodefplural{GP}[GPs]{Gaussian Processes}

\acro{KF}{Kalman Filter}

\acro{IP}{Interior Point}
\acro{BO}{Bayesian Optimization}

\acro{SE}{squared exponential}

\acro{UI}{uncertain input}

\acro{MCL}{Monte Carlo Localisation}
\acro{AMCL}{Adaptive Monte Carlo Localisation}

\acro{SSIM}{Structural Similarity Index}

\acro{MAE}{Mean Absolute Error}
\acro{RMSE}{Root Mean Squared Error}
\acro{AUSE}{Area Under the Sparsification Error curve}

\acro{AL}{active learning}
\acro{DL}{deep learning}
\acro{CNN}{convolutional neural network}

\acro{MC}{Monte-Carlo}

\acro{GSD}{ground sample distance}

\acro{BALD}{Bayesian active learning by disagreement}

\acro{fCNN}{fully convolutional neural network}
\acro{FCN}{fully convolutional neural network}

\acro{CMA-ES}{Covariance Matrix Adaptation Evolution Strategy}

\acro{FoV}{field of view}

\acro{mIoU}{mean Intersection-over-Union}
\acro{ECE}{expected calibration error}

\end{acronym}

\IEEEoverridecommandlockouts                           

\title{NeU-NBV: Next Best View Planning Using Uncertainty Estimation\\ in Image-Based Neural Rendering}

\author{Liren Jin, Xieyuanli Chen, Julius R\"{u}ckin, Marija Popovi\'{c}
\thanks{This work has been fully funded by the Deutsche Forschungsgemeinschaft (DFG, German Research Foundation) under Germany's Excellence Strategy, EXC-2070 -- 390732324 (PhenoRob) and supported by the NVIDIA Academic Hardware Grant Program. All authors are with the Institute of Geodesy and Geoinformation, University of Bonn.
Corresponding: \texttt{ljin@uni-bonn.de}.}
}

\begin{document}

\maketitle

\begin{abstract}
Autonomous robotic tasks require actively perceiving the environment to achieve application-specific goals. In this paper, we address the problem of positioning an RGB camera to collect the most informative images to represent an unknown scene, given a limited measurement budget. We propose a novel mapless planning framework to iteratively plan the next best camera view based on collected image measurements. A key aspect of our approach is a new technique for uncertainty estimation in image-based neural rendering, which guides measurement acquisition at the most uncertain view among view candidates, thus maximising the information value during data collection. By incrementally adding new measurements into our image collection, our approach efficiently explores an unknown scene in a mapless manner.
We show that our uncertainty estimation is generalisable and valuable for view planning in unknown scenes. Our planning experiments using synthetic and real-world data verify that our uncertainty-guided approach finds informative images leading to more accurate scene representations when compared against baselines.
\end{abstract} 

\section{Introduction} \label{S:intro}
Active perception and exploration is a core prerequisite for embodied robotic intelligence. 
In many applications, including robotic manipulation, inspection, and vision-based navigation, 
the ability to autonomously collect data is crucial for scene understanding and further downstream tasks~\citep{Chen2011}.
A key challenge in this procedure is planning a view sequence for sensors to obtain the most useful information given platform-specific constraints~\citep{Bajcsy2018}.

In this work, we present a new framework for iteratively planning the next best view (NBV) for an RGB camera to explore an unknown scene. Given a limited measurement budget, our goal is to actively position the sensor to gather the most informative data of the scene online, i.e. during a robotic mission. To address this problem, traditional NBV planning methods rely on explicit global map representations of the scene, such as point clouds~\citep{Rui2020}, volumes~\citep{Isler2016, Zaenker2021, Bircher2016}, or surface meshes~\citep{Guillaume2020}, as a basis for planning. 
However, due to map discretisation, these approaches scale poorly to larger scenes and have limited representation ability~\citep{Tewari2022}.
\begin{figure}[!t]
\centering
  \begin{subfigure}[]{\columnwidth}
  \includegraphics[width=\columnwidth]{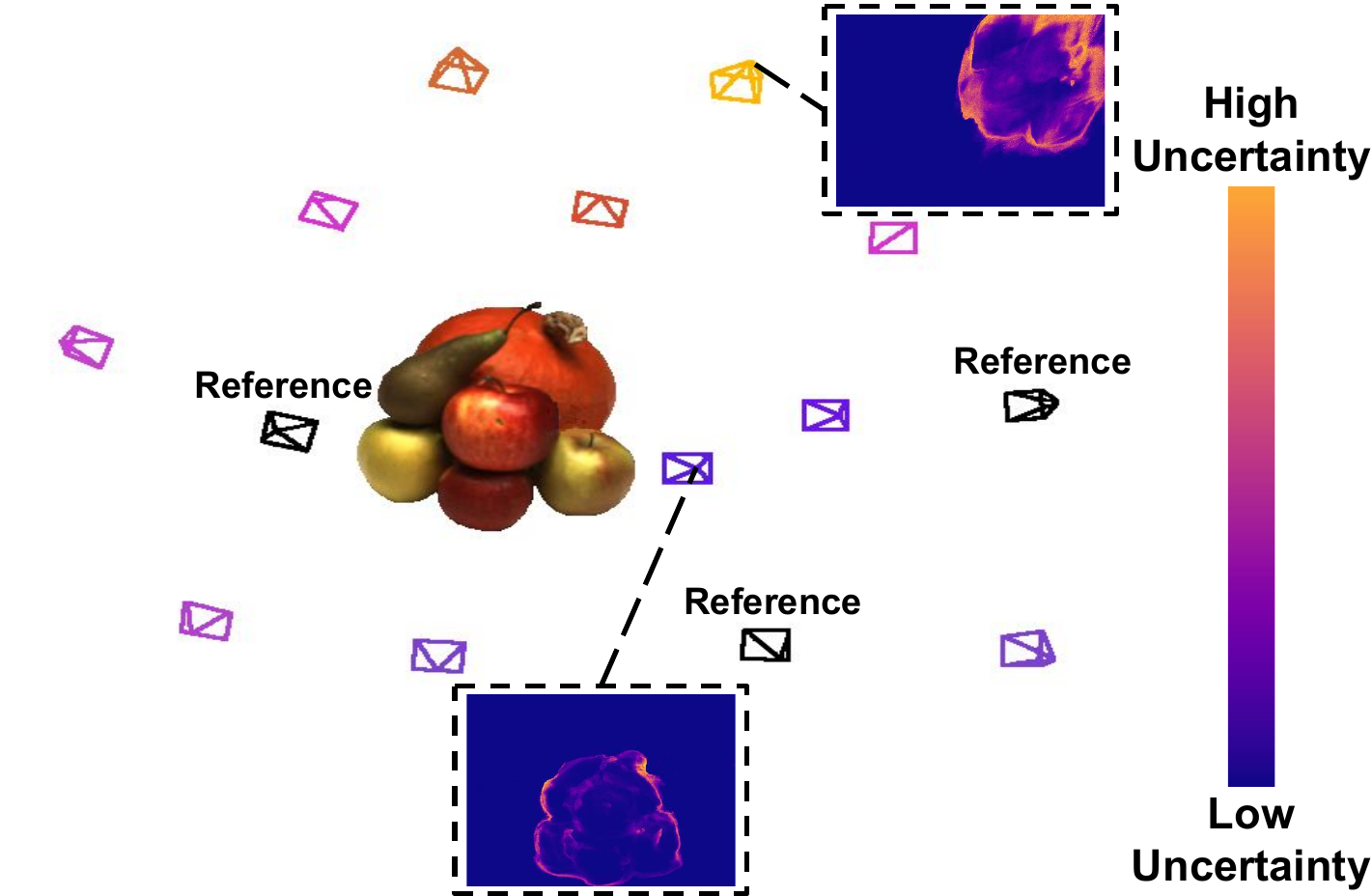}
  \end{subfigure}
    \caption{Our novel NBV planning framework exploits uncertainty estimation in image-based neural rendering to guide measurement acquisition. Given reference images from the current image collection of the scene (black frustums), our network outputs per-pixel uncertainty estimates at sampled view candidates (coloured frustums). Brighter frustums indicate higher average uncertainty from the view. Zoom-in boxes illustrate per-pixel uncertainty estimates at the most certain and uncertain views.
    By selecting the most informative, i.e. most uncertain, view candidate at which to take the next measurement, our approach efficiently explores the unknown scene in a mapless manner.}\label{F:teaser}
\vspace{-0.4cm}
\end{figure}

Implicit neural representations, such as neural radiance fields (NeRFs)~\citep{Mildenhall2020}, are drawing immense interest as an alternative approach for complex scene understanding. Given posed 2D images, NeRFs synthesise novel views by optimising an underlying continuous function encoding the scene appearance and geometry. In the context of active perception, emerging works~\citep{Pan2022, Zhan2022, Ran2023, Sunderhauf2022, Lee2022} incorporate uncertainty estimation into NeRFs and exploit it to guide NBV planning. While showing promising results, these studies follow an active learning~\citep{ren2021} paradigm to collect the most informative, i.e. most uncertain, images for periodically re-training a NeRF to improve the scene representation with minimal data. Since re-training NeRF models is computationally expensive, such methods are impractical for online robotic applications.

To overcome the inefficiency and per-scene optimisation requirements of NeRF models, another line of work focuses on image-based neural rendering~\citep{Yu2021, Rosu2021, Wang2021, Trevithick2021}. Image-based approaches exploit a shared encoder to map nearby 2D reference images into latent feature space, upon which the local implicit representation is conditioned. This allows training the network across multiple scenes to learn scene priors, enabling it to generalise to new scenes without test time optimisation. 
Previous works in image-based neural rendering~\citep{Yu2021, Rosu2021, Wang2021, Trevithick2021} mainly study improving network performance assuming pre-recorded image data. However, exploiting the benefits of image-based neural rendering for active view planning in robotics has not yet been considered.

The main contribution of this paper is a novel NBV planning framework bridging the gap between active perception and image-based neural rendering for online robotic applications.
A key aspect of our framework is a new technique for uncertainty estimation in image-based neural rendering, which enables us to quantify the informativeness of view candidates without relying on ground truth images or global scene representations. Intuitively, high uncertainty indicates where scene information provided by the closest reference images is insufficient to render the novel view, due to under-sampling or more complex scene details in these areas. Therefore, we utilise view uncertainty as an informative exploration objective.
As shown in~\cref{F:teaser}, based on the predicted uncertainty, we actively select the most uncertain view candidate to maximise the information acquired during a data collection process in an unknown scene. 
 
Our work addresses active perception using learned uncertainty in image-based neural rendering. We make the following three claims: (i)~our uncertainty estimation technique generalises to unknown scenes and provides an informative proxy for rendering quality at novel views; (ii)~our uncertainty-guided NBV planning strategy outperforms baseline approaches in finding more informative images to represent an unknown scene; 
(iii)~the informative images collected using our approach also improve the offline training quality of NeRF models. 
To support reproducibility, our implementation and simulation dataset will be released at: \url{https://github.com/dmar-bonn/neu-nbv}.

\section{Related Work} \label{S:related_work}
\subsection{Next Best View Planning} \label{SS:next_best_view}
View planning for robot active perception is an area of active research~\citep{Bajcsy2018}.
In initially unknown scenes, a common approach is to iteratively select the NBV from a set of view candidates using an acquisition function capturing their expected utility based on the current map state.

\citet{Isler2016} build a probabilistic volumetric map and select the NBV by calculating the information gain composed of visibility and the likelihood of seeing new parts of an object. \citet{Bircher2016} find the NBV in a receding-horizon fashion by generating a random tree and selecting the branch maximising the amount of unmapped space in a volumetric map from view candidates. \citet{Zaenker2021} maintain a voxel map of the scene and select the NBV among candidates obtained by targeted region-of-interest sampling and frontier-based exploration sampling. \citet{Rui2020} propose a point cloud-based deep neural network to directly predict the information gain of view candidates from the current raw point cloud of the scene. \citet{Song2018} evaluate the completeness of reconstructed surfaces and extract low-confidence surfaces to guide NBV planning. 
All these approaches require explicit discretised 3D map representations to maintain current information about the scene, which limits their scalability and representation ability.
In contrast, our approach utilises a compact implicit neural representation conditioned only on 2D image inputs for NBV planning.

\subsection{Implicit Neural Representations} 
\label{SS:neural_representation}
Implicit neural representations parameterise a continuous differentiable signal with a neural network~\citep{Tewari2022}.
For example, NeRFs~\citep{Mildenhall2020} learn a density and radiance field supervised only by 2D images. To render a novel view, NeRFs sample points densely along a camera ray, then predict radiance and density from the position and view direction of each point. The final RGB and depth estimate of the ray is calculated by differentiable volume rendering. As the scene information is encoded in the network parameters, NeRFs overfit to a single scene and require significant training time. 

Instead of memorising a specific scene, image-based neural rendering, e.g. PixelNeRF~\citep{Yu2021}, leverages an encoder to map nearby reference images into latent feature space. After aggregating features from reference images, a multilayer perceptron (MLP) is trained to interpret the aggregated feature into appearance and geometry information at a novel view. By training across different scenes, image-based approaches generalise well to new scenes without test time optimisation. We exploit the generalisation ability of image-based neural rendering to achieve online NBV planning for efficient data collection in an unknown scene.

\subsection{Uncertainty Estimation in Neural Representations} 
\label{SS:neural_uncertainty}
Estimating uncertainty in learning-based computer vision tasks is a long-standing problem~\citep{Kendal17}.
Several recent works address uncertainty quantification in NeRF models. S-NeRF~\citep{Shen2021} proposes learning a probability distribution over all possible radiance fields modelling the scene. To this end, it treats radiance and density as stochastic variables and uses variational inference to approximate their posterior distribution after training. W-NeRF~\citep{martinbrualla2021} directly learns to predict RGB variance as an uncertainty measure in rendering transient objects in the scene. For image-based neural rendering, \citet{Rosu2021} introduce a loss function to learn confidence estimation in the rendered images. However, they only consider a fixed number of reference images with small view changes as inputs, which limits the applicability of their approach in robotics.
A contemporary work~\citep{Smith2022} leverages the occupancy predictions to estimate uncertainty for active vision. Their approach mainly handles single object shape reconstruction and requires known foreground masks.

Emerging works use uncertainty-guided NBV selection to address NeRF training with a constrained measurement budget. \citet{Pan2022} and \citet{Ran2023} model the emitted radiance as Gaussian distribution and learn to predict the variance by minimising negative log-likelihood during training. These works add the view candidate with the highest information gain, i.e. the highest uncertainty reduction, to the existing training data. Instead of learning uncertainty in parallel to radiance and density, \citet{Lee2022} and \citet{Zhan2022} propose calculating the entropy of the density prediction along the ray as an uncertainty measure with respect to the scene geometry. The entropy is used to guide measurement acquisition towards less precise parts. \citet{Sunderhauf2022} exploit the recent development of fast rendering of Instant-NGP~\citep{mueller2022} to train an ensemble of NeRF models for a single scene, and measure uncertainty using the variance of the ensemble's prediction, which is utilised for NBV selection.

The above-mentioned approaches address uncertainty-guided NBV selection based on NeRFs. Although these approaches show NeRF model refinement with limited input data, deploying such methods in robotic applications is not straightforward. 
As the scene information is entirely encoded in the network weights, after each planning step, the uncertainty estimation must be re-optimised to account for newly added measurements, which is time- and compute-consuming. In contrast, our novel approach incorporates uncertainty estimation in image-based neural rendering to actively select informative images, which are incrementally added to our image collection. This way, we explore an unknown scene without the need to maintain an explicit map representation or re-train an implicit neural representation.
 
\section{Our Approach} \label{S:approach}
\begin{figure}[t]
\centering
  \begin{subfigure}[]{\columnwidth}
  \includegraphics[width=\columnwidth]{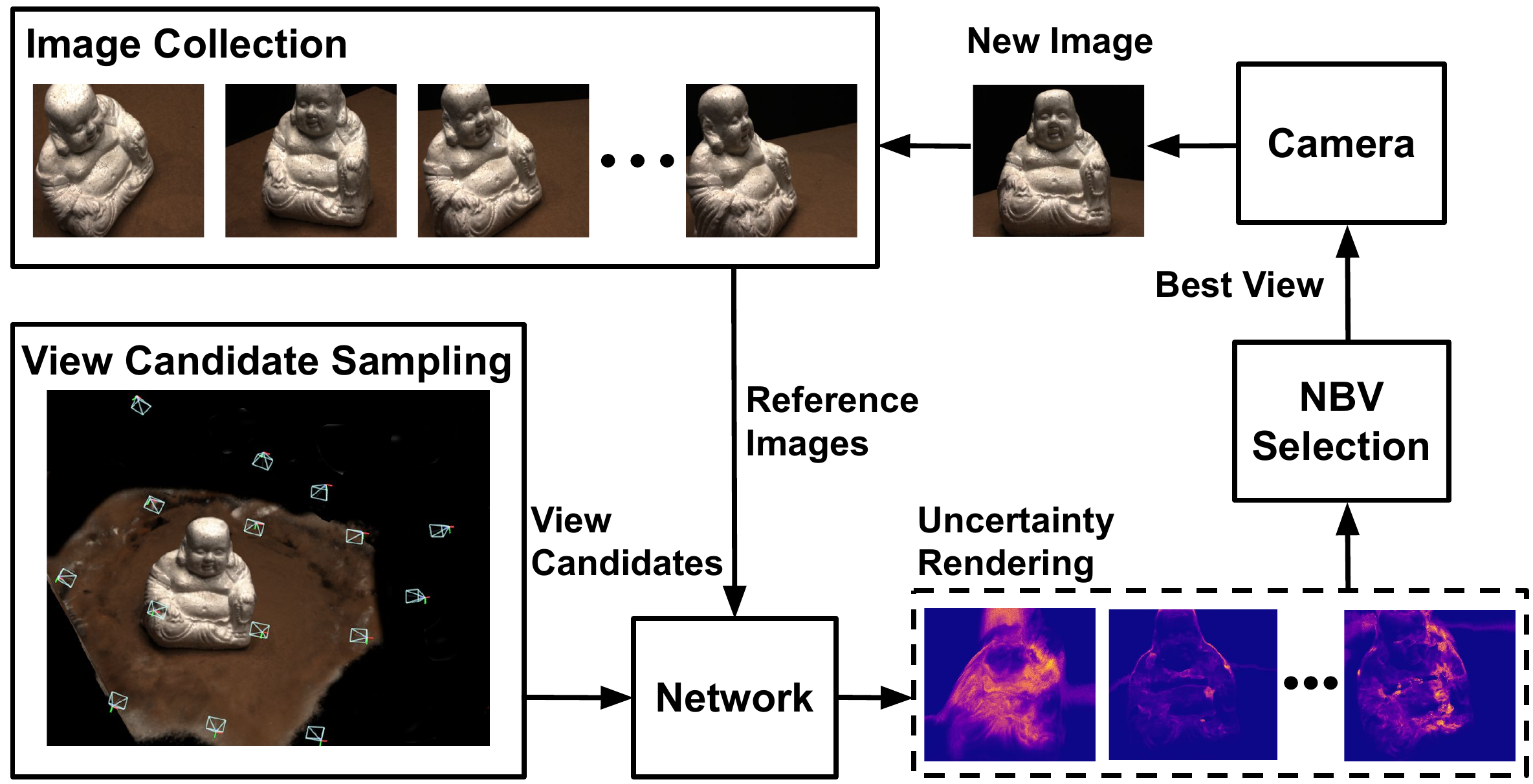}
  \end{subfigure}
    \caption{Overview of our mapless NBV planning framework. We leverage uncertainty estimation in image-based neural rendering to actively guide measurement acquisition in unknown scenes.}\label{F:framwork}
\vspace{-0.3cm}
\end{figure}
We propose a novel mapless NBV planning framework for robotic exploration tasks. An overview of our framework is shown in~\cref{F:framwork}. We first sample view candidates and query their corresponding closest reference images from the current image collection. Based on the scene information provided by the reference images, our image-based neural rendering network predicts per-pixel uncertainty at these view candidates. The NBV planning strategy selects the most uncertain view candidate corresponding to the next measurement, which we add to the image collection. Our image-based neural rendering network retrieves scene information in a purely image-based manner. This enables us to achieve efficient autonomous exploration without maintaining an explicit map or iteratively re-training an implicit neural representation. In the following subsections, we describe our network architecture, training procedure for uncertainty estimation, and NBV planning scheme.
\subsection{Network Architecture} \label{SS:network_architecture}
\begin{figure*}[t]
\centering
  \includegraphics[width=0.9\textwidth]{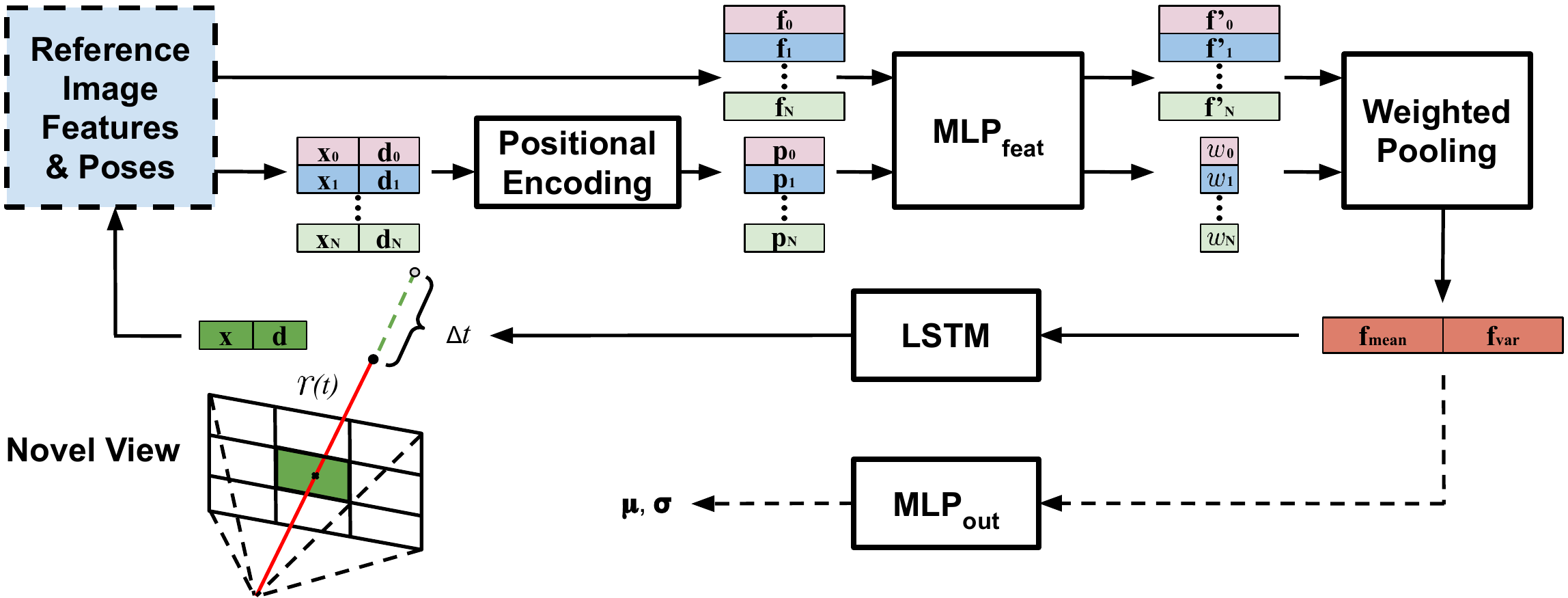}
  \caption{Our network architecture. Different colours indicate features from different reference images. Note that the encoder is not explicitly shown. We use an LSTM module to predict jumping distance $\Delta t$ to the next sampling point given the aggregated feature from all reference images acquired at the current sampling point. After a fixed number of iterations, the aggregated feature at the final point is interpreted to colour and uncertainty information. Arrows with dashed lines show the forward pass happening only in the last iteration.}
  \label{F: network}
\vspace{-0.3cm}
\end{figure*}
Our network follows the design choices of PixelNeRF~\citep{Yu2021} regarding the architecture of encoder and MLP. However, PixelNeRF uses a volume rendering technique requiring dense sampling along the ray at predefined intervals, which is inefficient and limits its online applicability. Inspired by~\citet{Rosu2021} and \citet{Sitzmann2019}, we adopt a long short-term memory (LSTM) module~\citep{Hochreiter1997} to adaptively predict the jumping distance to the next sampling point, therefore speeding up the inference of neural rendering. The network is illustrated in~\cref{F: network}.

 Given a novel view, we query our current image collection to find the $N$ closest reference images $\mathbf{I}_{n \in \{1, 2,\ldots,N\}}$. We use a shared convolutional-based encoder $E$ to extract latent feature volume $\mathbf{F}_n = E(\mathbf{I}_n)\in \mathbb{R}^{H \times W \times L}$ from each reference image. $H$ and $W$ are feature volume's spatial resolution and $L$ is the channel dimension. We parameterise a ray emitted from the novel view as $r(t) = \mathbf{o} + t \mathbf{d}$, where $\mathbf{o} \in \mathbb{R}^3$ is the camera centre position and $t$ is the distance along view direction $\mathbf{d} \in \mathbb{R}^3$.
 Starting from the close end of the ray \mbox{$t = t_s$}, we transform the sampling point's position $\mathbf{x} = r(t)$ and view direction $\mathbf{d}$ into each reference view coordinate using known relative camera poses to get $\mathbf{x}_n$ and $\mathbf{d}_n$, respectively. To recover high-frequency details of the scene, the point position $\mathbf{x}_n$ is mapped into higher-dimensional space by the positional encoding operation $\gamma$ proposed by~\citet{Mildenhall2020}. By combining it with its view direction, we compose pose feature $\mathbf{p}_n = (\gamma(\mathbf{x}_n), \mathbf{d}_n)$ for sampling point expressed in $n^{th}$ reference view coordinate. 
To retrieve the latent image feature from reference images, we project $\mathbf{x}_n$ onto the corresponding reference image plane using known camera intrinsics to get image coordinates $\phi(\mathbf{x}_n)$, which we use to query the image feature $\mathbf{f}_n = \mathbf{F}_n(\phi(\mathbf{x}_n))\in \mathbb{R}^{L}$ by grid sampling with bilinear interpolation~\citep{Yu2021}.

The acquired pose feature $\mathbf{p}_n$ and image feature $\mathbf{f}_n$ from each reference image are processed individually by MLP\textsubscript{feat}.
For aggregating features from all 
reference images, we use the predicted weight $w_n \in \left [ 0\,,1\right]$ and processed feature $\mathbf{f}_n^{\prime}$ to calculate the weighted mean $\mathbf{f}_\mathrm{mean}$ and variance $\mathbf{f}_\mathrm{var}$.
This operation downweights the feature from less informative reference images. Conditioning on the aggregated feature $(\mathbf{f}_\mathrm{mean}, \mathbf{f}_\mathrm{var})$, our LSTM module adaptively predicts the jumping distance $\Delta t$ to the next sampling point $\mathbf{x} = r(t+\Delta t)$, thus mitigating the sampling inefficiency commonly seen in volume rendering~\citep{Yu2021, Mildenhall2020}. We iterate this process a fixed number of times to let the sampling point approach the surface in the scene and acquire depth prediction. We then use MLP\textsubscript{out} to interpret the aggregated feature queried at the final sampling point into colour and uncertainty information, as detailed in the following subsection.

\subsection{Uncertainty Estimation in Image-based Neural Rendering} \label{SS:uncertainty_estimation}
Our uncertainty estimation quantifies the uncertainty inherited from the input data, due to the varying quality of the information provided by the reference images. For example, we expect reference images with large view differences and self-occlusions with respect to the novel view to lead to blurry rendering and thus high uncertainty. An illustration of input-dependent uncertainty estimated using our new approach is shown in~\cref{F:rendering_visual}.

Given supervision using only posed 2D images, we incorporate input-dependent uncertainty estimation in the image-based neural rendering training process.
Considering that the predicted RGB value is normalised between $\left [ 0\,,1\right]$, we model each channel value of the RGB prediction $c_{i} \in \left [ 0\,,1\right]$, where $i \in \{1, 2, 3\}$, as an independent logistic normal distribution described by: 
\begin{align}
     p(c_{i}; \mu_{i}, \sigma_{i}) = \frac{1}{\sigma_{i}\sqrt{2\pi }} \, \frac{1}{c_{i}(1-c_{i})} \, e^{-\frac{(\mathrm{logit}(c_{i}) - \mu_{i})^2}{2\sigma_{i}^2}}\,,
     \label{E:color_variable}
\end{align}
where $\mathrm{logit}(c_{i}) = \mathrm{ln}(\frac{c_{i}}{1-c_{i}}) \sim \mathcal{N}(\mu_{i},\,\sigma_{i}^{2}) $ follows a normal distribution, with the mean $\mu_{i}$ and variance $\sigma_{i}^{2}$ predicted by our network. To train the network, following~\citet{Kendal17}, we minimise the negative log-likelihood $-\log p\left (c_{i} = y_{i}\mid\mu_{i}, \sigma_{i}\right)$ given ground truth RGB channel values $y_{i} \in \left [ 0\,,1\right]$. For a single pixel RGB prediction, this leads to our photometric loss function formulated as:
\begin{equation}
    \small{
     \mathcal{L} = \sum_{i=1}^{3}\frac{1}{2} \log(\sigma_{i}^2) + \log(y_{i}(1-y_{i})) + \frac{(\mathrm{logit}(y_{i}) - \mu_{i})^2}{2\sigma_{i}^2}\,.
    }
     \label{E:loss_function}
\end{equation}
For calculating the loss, the ground truth RGB channel value is mapped into logit space by $\mathrm{logit}(y_{i})$. We clamp $y_{i}$ at $\left [ 0.001\,, 0.999\right]$ to ensure numerical stability. 

During deployment in unknown scenes, given a novel view and its reference images, our network predicts mean $\mu_i$ and variance $\sigma_i^2$ assuming each RGB channel of a pixel is normally distributed in logit space. We sample $100$ times from the normal distribution and pass all samples through a sigmoid function to acquire a valid RGB channel value. The mean and variance of the $100$ channel values represents our final channel-wise RGB prediction $c_{i} \in \left [ 0\,,1 \right ]$ and the corresponding uncertainty estimate $u_{i} \in \left [ 0\,,0.25 \right ]$.

\subsection{Uncertainty Guided Next Best View Planning} \label{SS: planning_strategy}
Our novel NBV planning framework exploits uncertainty estimation in image-based neural rendering to guide efficient data collection. Given a limited measurement budget, our uncertainty-guided approach is effective at finding more informative images to better represent an unknown scene.

For view planning, we consider a scene-centric hemisphere action space. First, our planning procedure initialises the image collection with image measurements at two random views. For planning the next camera view, we uniformly sample a fixed number of view candidates $k \in \{1, 2,\ldots,K\}$ within allowable view changes around the current camera view. For each view candidate, we find at maximum $N$ closest reference images in our current image collection. Given the novel view and corresponding reference images, our network renders per-pixel uncertainty estimate $\mathbf{U}_{k} \in \left [ 0\,,0.25 \right ]^{H_r \times W_r \times 3}$ following the approach in~\cref{SS:neural_uncertainty}, where $H_r$ and $W_r$ are the desired rendering resolution. 
In this setup, we propose a simple utility function defined as:
\begin{equation}
    g(k) = \frac{1}{H_r \times W_r \times 3} \,  \mathrm{sum}(\mathbf{U}_{k}). 
\end{equation}
The view candidate $k^{*}$ with the highest utility $g(k^{*})$ is selected as our NBV. High uncertainty indicates that the view candidate cannot be well-rendered by our network given the current image collection, due to under-sampling around the view, i.e. the closest reference images are far away, or the scene is generally complex when observed from the view. Therefore, a new measurement at the most uncertain view potentially yields the highest information value for scene representation. The newly-captured image at the NBV is then added to our image collection. We iterate this planning procedure until a given measurement budget is exhausted. 
Note that our framework is agnostic to sampling strategies and can be easily adapted to other specific scenarios.

\section{Experimental Evaluation} 
\begin{figure*}[!t]
\centering
\begin{subfigure}{0.9\columnwidth}
\includegraphics[width=\columnwidth]{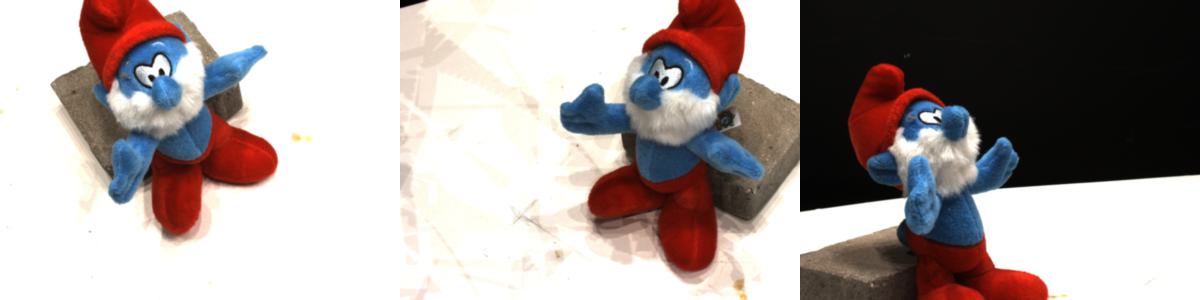}%
  \caption*{Reference image set 1} \label{SF:mapping_results_gt}
\end{subfigure}\hspace{8mm}
\begin{subfigure}{.3\columnwidth}
\includegraphics[width=\columnwidth]{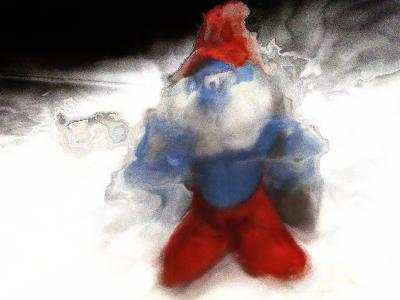}%
  \caption*{}
\end{subfigure}
\begin{subfigure}{.3\columnwidth}
\includegraphics[width=\columnwidth]{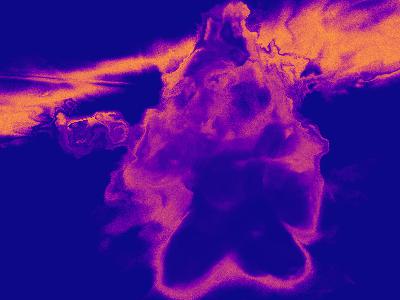}%
  \caption*{}
\end{subfigure}
\begin{subfigure}{.3\columnwidth}
\includegraphics[width=\columnwidth]{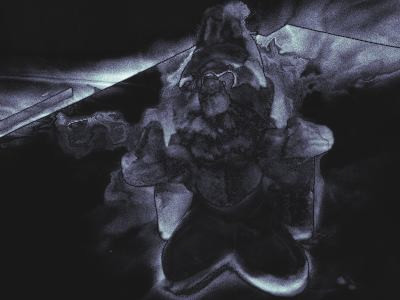}%
  \caption*{}
\end{subfigure}

\begin{subfigure}{0.9\columnwidth}
\includegraphics[width=\columnwidth]{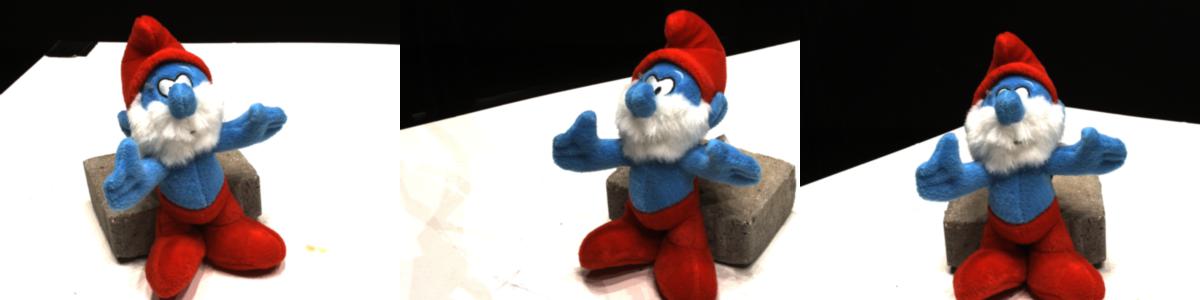}%
  \caption*{Reference image set 2}
\end{subfigure}\hspace{8mm}
\begin{subfigure}{.3\columnwidth}
\includegraphics[width=\columnwidth]{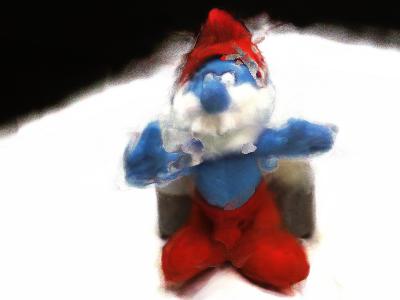}%
  \caption*{Rendered RGB}
\end{subfigure}
\begin{subfigure}{.3\columnwidth}
\includegraphics[width=\columnwidth]{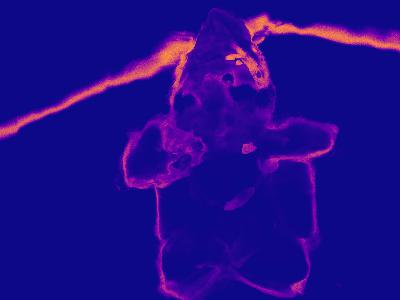}%
  \caption*{Uncertainty}
\end{subfigure}
\begin{subfigure}{.3\columnwidth}
\includegraphics[width=\columnwidth]{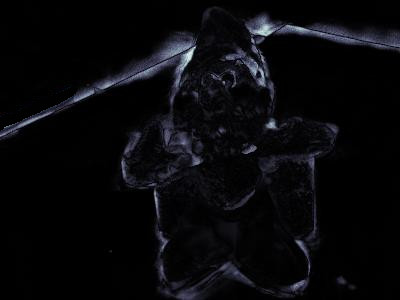}%
  \caption*{Error}
\end{subfigure}
\caption{Examples of our input-dependent uncertainty estimation in image-based neural rendering. For rendering the same novel view, we select two sets of reference images. The comparison clearly shows how rendering quality depends on scene information provided by reference images. 
Reference images with low information value lead to blurry rendering and correspondingly high uncertainty prediction (yellower) from our network. The error map shows the mean squared error between ground truth and rendered RGB (whiter areas indicate higher errors). Our uncertainty prediction is strongly correlated with this error, thus serving as a good proxy for view planning.
}\label{F:rendering_visual}
\end{figure*}
\begin{table*}[!h]
\caption{Evaluation of uncertainty estimation strategies on the 15 test scenes of the DTU dataset. Best results in bold.}
\label{T:calibration_table}\textbf{}
\centering
\setlength{\tabcolsep}{5pt}
\begin{tabular}{@{}ccccccccccccccccc@{}}
\toprule
Scene No. && 8 & 21 & 30 & 31 & 34 & 38 & 40 & 41 & 45 & 55 & 63 & 82 & 103 & 110 & 114\\ \midrule
 \multirow{3}{*}{SRCC $\uparrow$} 
  &Entropy      & $0.16$   & $0.52$   & $0.37$ & $0.29$ & $0.21$ & $0.60$ & $0.39$  & $0.52$  & $0.17$  & $0.47$ & $0.53$ & $0.32$ & $0.42$ & $0.33$ & $0.60$ \\
  &Confidence   & $0.83$   & $0.83$   & $0.90$ & $0.80$ & $0.66$ & $0.76$ & $0.81$  & $0.80$  & $0.83$  & $0.78$ & $0.82$ & $0.88$ & $0.48$ & $0.53$ & $0.79$ \\
  &Ours         & $\mathbf{0.84}$   & $\mathbf{0.89}$   & $\mathbf{0.93}$ & $\mathbf{0.88}$ & $\mathbf{0.86}$ & $\mathbf{0.87}$ & $\mathbf{0.83}$  & $\mathbf{0.86}$  & $\mathbf{0.89}$  & $\mathbf{0.91}$ & $\mathbf{0.91}$ & $\mathbf{0.93}$ & $\mathbf{0.73}$ & $\mathbf{0.83}$ & $\mathbf{0.89}$ \\ \midrule
 \multirow{3}{*}{AUSE $\downarrow$} 
  & Entropy     & $0.50$   & $0.48$   & $0.34$ & $0.42$ & $0.55$ & $0.48$ & $0.50$  & $0.51$  & $0.51$  & $0.41$ & $0.38$ & $0.34$ & $0.47$ & $0.36$ & $0.45$ \\
  &Confidence   & $0.25$   & $0.26$   & $0.14$ & $0.18$ & $0.21$ & $0.28$ & $0.27$  & $0.22$  & $0.19$  & $0.23$ & $0.14$ & $0.16$ & $0.23$ & $0.20$ & $0.16$ \\
  &Ours         & $\mathbf{0.17}$   & $\mathbf{0.18}$   & $\mathbf{0.05}$ & $\mathbf{0.11}$ & $\mathbf{0.12}$ & $\mathbf{0.19}$ & $\mathbf{0.14}$  & $\mathbf{0.13}$  & $\mathbf{0.11}$  & $\mathbf{0.15}$ & $\mathbf{0.08}$ & $\mathbf{0.08}$ & $\mathbf{0.18}$ & $\mathbf{0.12}$ & $\mathbf{0.11}$ \\
 \bottomrule
\end{tabular}
\vspace{-0.1cm}
\end{table*}

Our experimental results support our three claims: (i) we show that our uncertainty estimation in image-based neural rendering is informative to rendering quality and generalises to new scenes; (ii) we show that our uncertainty-guided NBV planning strategy collects informative images using a publicly available real-world dataset and in a simulated environment. To measure the quality of collected images, we evaluate their influence on image-based neural rendering performance at test views; and (iii) we show the benefit of using our online collected images to train NeRF models. Experimental results indicate that images collected using our planning framework lead to more accurate implicit representations in both cases when compared against baselines. 

\subsection{Training Procedure} \label{SS:training}

\textbf{Datasets.} We train our network separately on two datasets for the corresponding planning experiments. We first use real-world images with a resolution of $400 \times 300$ pixels from the DTU dataset~\citep{Rasmus2014}. We follow the data split proposed by PixelNeRF~\citep{Yu2021} with $88$ training scenes and $15$ test scenes, in which no shared or similar scenes exist.
For each scene, $49$ images are collected following a fixed path pattern on a section of a scene-centric hemisphere.
We also record our own synthetic dataset, considering $50$ ShapeNet~\citep{Chang2015} models from $4$ representative categories: car, motorcycle, camera, and ship. For each model, we record $100$ images with a resolution of $200 \times 200$ pixels from views uniformly distributed on the hemisphere covering the scene.

\textbf{Training Setup.} We use the Adam optimiser with a learning rate of $10^{- 5}$ and exponential decay of $0.999$. LSTM iteration number during a forward pass is set to $16$. The network is implemented in PyTorch and trained with a single NVIDIA RTX A5000 GPU for $\sim2$ days until convergence. Rendering a novel view with the same resolution as the two dataset images takes $0.6$\,s and $0.3$\,s, respectively, which is $60$ times faster than PixelNeRF~\citep{Yu2021}. Our network design is agnostic to the number of input reference images. For both training processes, we randomly select $3$, $4$, or $5$ reference images for novel view rendering in the scene to restrict memory consumption.  

\subsection{Evaluation of Uncertainty Estimation} \label{SS:neural_uncertainty}
Our first experiment is designed to show that our uncertainty estimation strongly correlates with rendering error in image-based neural rendering in unknown scenes.
To evaluate the quality of uncertainty prediction, we consider two metrics. We use Spearman's Rank Correlation Coefficient (SRCC)~\citep{Spearman1904} to asses the monotonic relationship between averaged uncertainty estimate and rendering error over a test view. As SRCC only captures the informativeness of averaged uncertainty prediction, the quality with respect to the structural similarity between the per-pixel uncertainty estimate and error is not considered. To evaluate the structural similarity, we report the Area Under the Sparsification Error (AUSE) curve~\citep{Ilg2018}, which reveals how strongly the uncertainty coincides with the rendering error pixel-wise.

For each test scene in the DTU dataset, we create $100$ test sets. Each test set consists of four images randomly selected from the scene, from which we use three as reference images and the remaining one as the test view. We average the predicted uncertainty and mean squared error (MSE) of each test view. We then calculate SRCC values with respect to the $100$ pairs of averaged uncertainty and MSE. Empirically, SRCC values higher than $0.8$ indicate strong monotonicity (high average uncertainty prediction is consistent with high average rendering error). We also report the average AUSE over $100$ test views for each scene. AUSE of $0$ means that the order of pixel-wise uncertainty magnitude perfectly aligns with the order of the MSE value (uncertain areas at the rendered test view overlap with erroneous predictions). 

We compare our approach against two alternative uncertainty estimation methods that can be incorporated into image-based neural rendering frameworks. \citet{Lee2022} propose calculating the entropy of the density distribution of the samples along each ray as uncertainty quantification in NeRF models. We re-implement this entropy calculation in PixelNeRF, which we denote \textit{Entropy} in the experiments. \citet{Rosu2021} proposes learning to predict RGB rendering confidence in image-based neural rendering by defining the loss as a linear combination of the predicted and the ground truth images. As their approach only handles a fixed number of reference images with small view changes, we adapt it by replacing our loss function~\cref{E:loss_function} with their confidence loss and train the network under the same conditions. We denote this method \textit{Confidence}.

\cref{T:calibration_table} summarises the results. Our uncertainty prediction is more informative with respect to rendering error compared to the other two methods. The poor performance of the \textit{Entropy} approach is likely due to the fact that the entropy of the density distribution mainly captures uncertainty over scene geometry. As neural rendering can recover colour information under inaccurate depth prediction~\citep{Tewari2022}, naïvely incorporating \textit{Entropy} as uncertainty estimation in image-based neural rendering fails to provide useful information about rendering quality. The superior performance of our approach compared to \textit{Confidence} indicates that our probabilistic interpretation of RGB prediction leads to more consistent uncertainty estimates. A qualitative illustration of our uncertainty prediction results is exemplified in~\cref{F:rendering_visual}.

\subsection{Comparison of Next Best View Planning Strategies} \label{SS:NBV}
We show that our uncertainty-guided NBV planning collects the most informative images to better represent an unknown scene. For evaluating planning performance, we use collected images and our image-based neural rendering network to render test views. The rendering quality is measured by the peak signal-to-noise ratio (PSNR) and structural similarity index measure (SSIM)~\citep{Mildenhall2020}. Note that, since the image-based neural rendering network is fixed for test view rendering in all experiments, performance differences arise purely as the consequence of different NBV planning strategies.
We compare our uncertainty-guided approach against two heuristic baselines:
\begin{itemize}
\item \textit{Ours}: selects the most uncertain view candidate via our uncertainty prediction as illustrated in~\cref{F:teaser};
\item \textit{Max. View Distance}: selects the view candidate that maximises the view distance with respect to previously collected images;
\item \textit{Random}: selects a view candidate uniformly at random.
\end{itemize}

We conduct experiments on the DTU dataset and in our simulator with corresponding pre-trained networks, respectively. For all planning experiments, we initialise the image collection with two randomly selected images and use different planning approaches to take the next images until a given maximum of measurements is reached. 

\begin{figure}[t]
\centering
  \begin{subfigure}[]{\columnwidth}
  \includegraphics[width=\columnwidth]{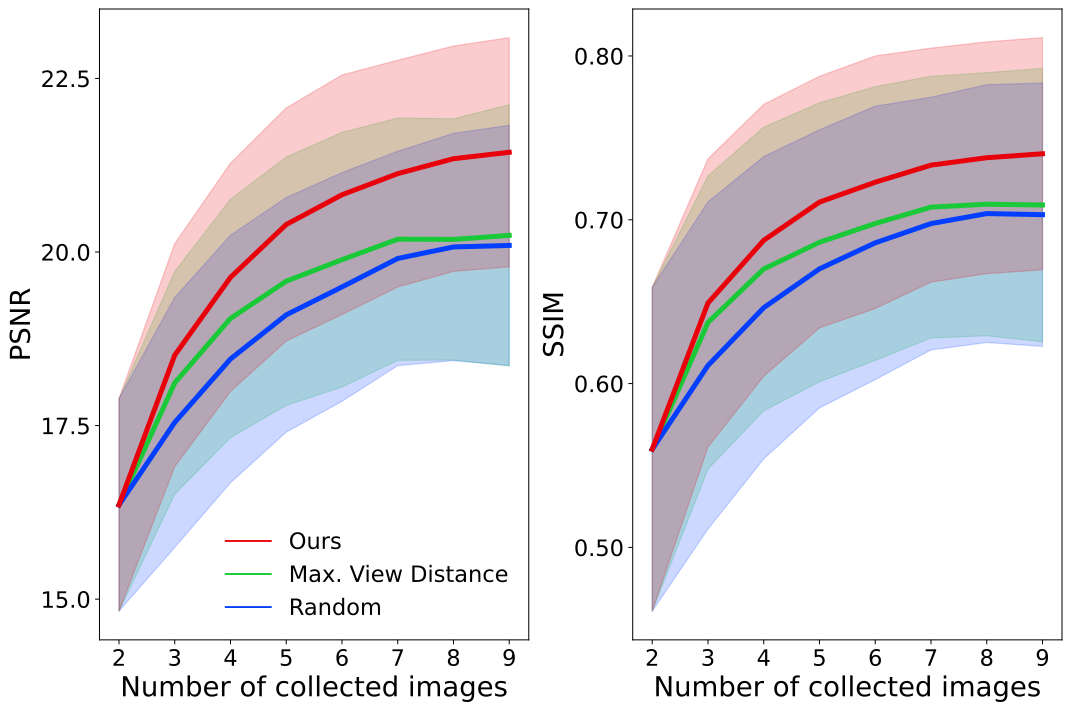}
  \end{subfigure}\hfill%
  \caption{Comparison of NBV planners on the DTU dataset. For each test scene, we use our image-based neural rendering network and collected images to render unselected views. To evaluate planning performance, we report the average PSNR and SSIM with standard deviations over all test scenes and runs. Note that the large standard deviations are due to varying rendering difficulty of each scene. Our uncertainty-guided approach finds informative images in the scene, improving scene representations via image-based neural rendering.}
  \label{F:dtu_planning}
  \vspace{-0.5cm}
\end{figure}

\textbf{DTU dataset.} 
We set the measurement budget to $9$ images including the $2$ images for initialisation. As the DTU dataset has limited views for each scene, we treat all unselected views as view candidates. We adopt the three planning strategies to select the next view from the view candidates and add it to our image collection. After each view selection step, we use the current image collection to render all unselected views. We calculate the average PSNR and SSIM with standard deviations. We repeat the experiment $10$ times for all $15$ test scenes and report the results in~\cref{F:dtu_planning}. As shown, NBV planning guided by our uncertainty estimation selects the most informative view candidate in each step reflected by better image-based neural rendering quality.
\begin{figure*}[!h]
\centering
\begin{subfigure}{\columnwidth}
  \includegraphics[width=\columnwidth]{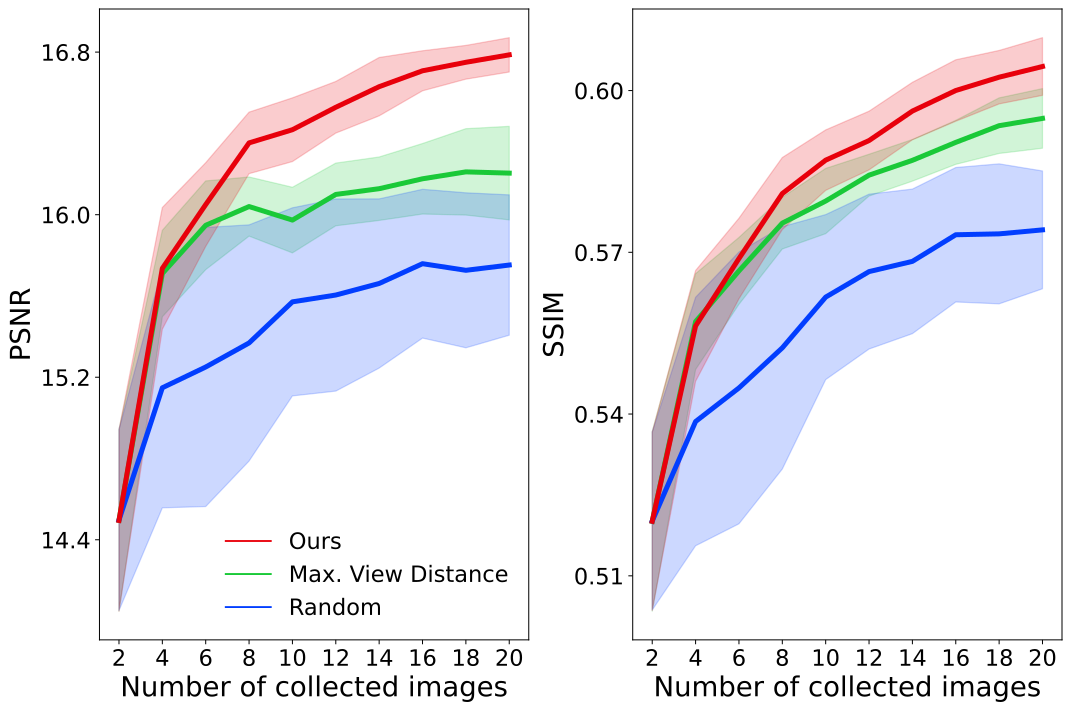}
  \caption{Car model}
\end{subfigure}
\begin{subfigure}{\columnwidth}
  \includegraphics[width=\columnwidth]{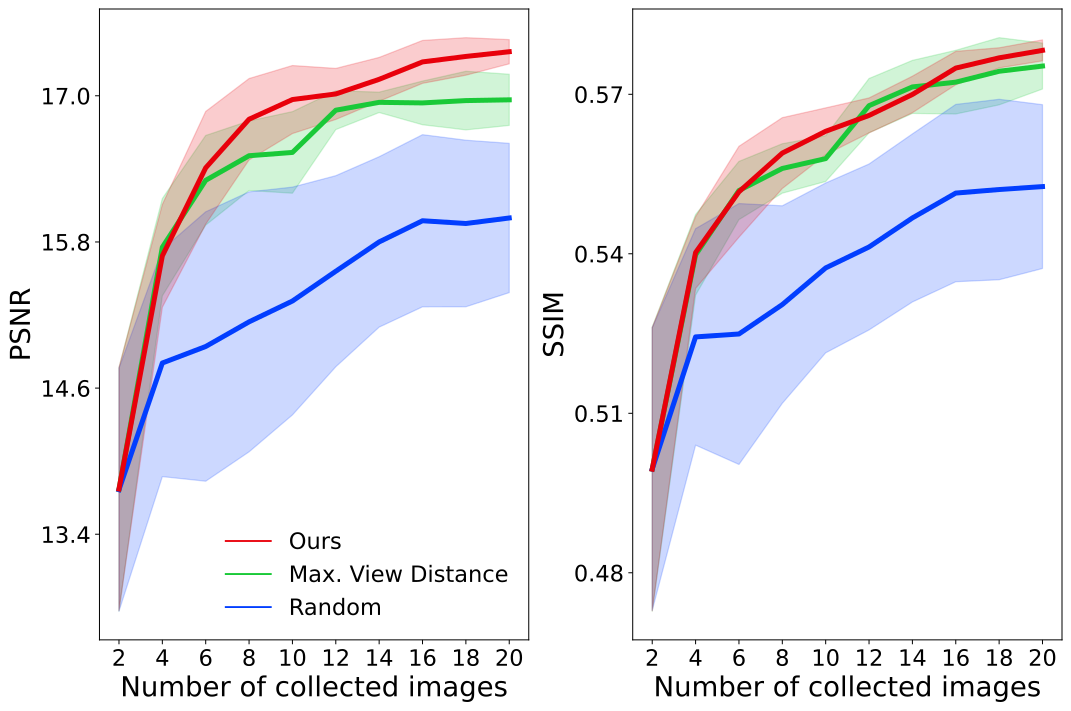}
  \caption{Indoor model}
\end{subfigure}
\caption{Comparison of NBV planners in a ShapeNet-based simulation environment. We conduct the experiments on (a) car and (b) indoor models, respectively. We fix $100$ test views for evaluation purposes. For each test view, we query a maximum of 5 closest reference images from currently collected images and use our image-based neural rendering network to render the test view. We report the average PSNR and SSIM with standard deviations over all test views and experiment runs. Our uncertainty-guided NBV planning outperforms heuristic baselines in finding more informative images, resulting in higher rendering quality given a limited measurement budget.}
\label{F: gazebo_planning}
\vspace{-0.3cm}
\end{figure*}

\textbf{Simulator.} To demonstrate the advantages of our NBV planning framework in a more realistic robotic application scenario, we show the planning experiment in a simulation environment with a continuous action space. We import two different ShapeNet 3D models into the simulator. First, we consider a car model, which belongs to the training category but is not seen during training. Second, to show the generalisation ability of our approach, we test our planning framework on an indoor model consisting of a sofa and table. Note that the sofa and table are not in our training data categories.
We configure our action space as a scene-centric hemisphere and set the measurement budget to $20$ images including $2$ initialisation images. At each planning step, we uniformly sample $50$ view candidates within the interval of maximal $60^\circ$ view angle change with respect to the current camera view. The three planners select the next view among the sampled view candidates. For our approach, we predict per-pixel uncertainty at $60 \times 60$ pixels resolution for each view candidate using a maximum of $5$ closest reference images. One planning step takes $1.5$\,s in this setting.
To evaluate the quality of collected images during online missions, we fix $100$ random test views of the scene. After every $2$ measurements, we use our network to render all test views given a maximum of $5$ closest reference images from the current image collection and report average PSNR and SSIM with standard deviations to evaluate implicit scene reconstruction quality. 
We repeat each planning experiment $10$ times on the two models, respectively.

\cref{F: gazebo_planning} summarises the planning results. Our findings confirm that images collected using our uncertainty-guided approach lead to better image-based neural rendering quality in both scenes. Non-adaptive heuristic approaches cannot efficiently utilise the measurement budget, thus limiting their view planning performance. In contrast, our uncertainty-guided approach collects informative images in a targeted way, resulting in higher test view rendering quality.

\subsection{Data Collection for Offline Modelling} \label{SS: offline} 
In this experiment, we further show that the images collected by our approach improve NeRF training using limited data. Note that different from uncertainty-guided NBV planning based on NeRFs~\citep{Pan2022, Ran2023, Sunderhauf2022, Lee2022}, our uncertainty estimation generalises to unknown scenes, thus the data collection process and NeRF training can be decoupled in our framework. This avoids computationally expensive network re-training during online missions.

\begin{figure*}[t]
\centering
\begin{subfigure}{0.45\columnwidth}
  \includegraphics[width=\columnwidth, height=2.7cm]{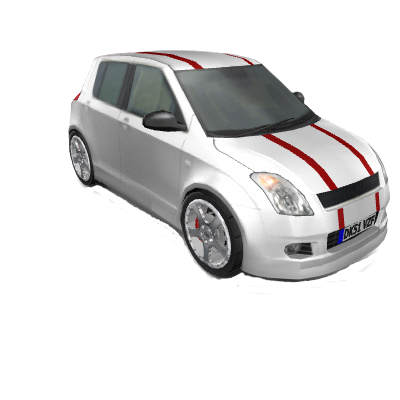}
\end{subfigure}
\begin{subfigure}{0.45\columnwidth}
  \includegraphics[width=\columnwidth, height=2.7cm]{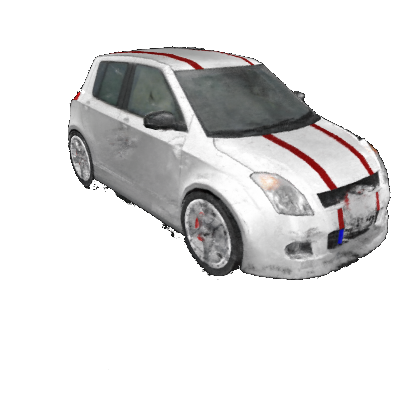}
\end{subfigure}
\begin{subfigure}{0.45\columnwidth}
  \includegraphics[width=\columnwidth, height=2.7cm]{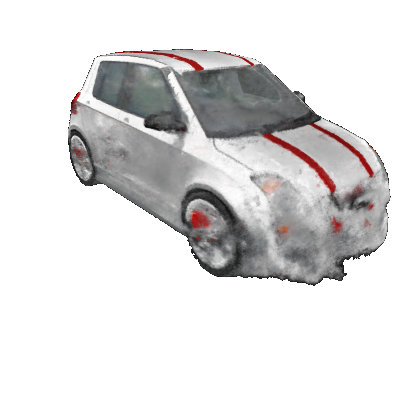}
\end{subfigure}
\begin{subfigure}{0.45\columnwidth}
  \includegraphics[width=\columnwidth, height=2.7
cm]{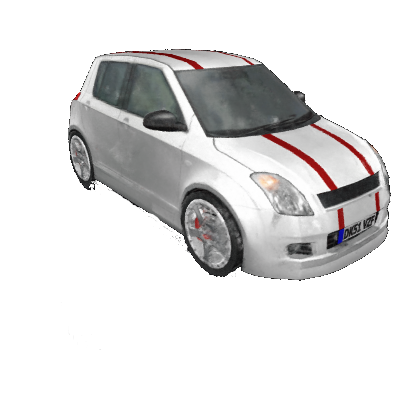}
\end{subfigure}

\vspace{-0.8cm}

\begin{subfigure}{0.45\columnwidth}
  \includegraphics[width=\columnwidth, height=2.7cm]{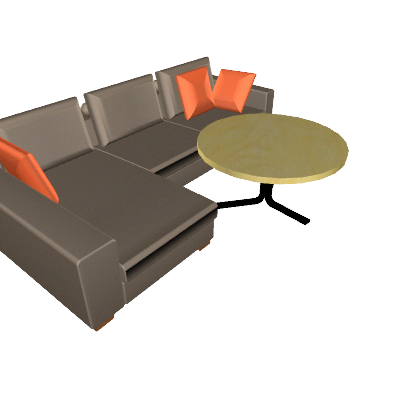}
  \caption*{Ground Truth}
\end{subfigure}
\begin{subfigure}{0.45\columnwidth}
  \includegraphics[width=\columnwidth, height=2.7cm]{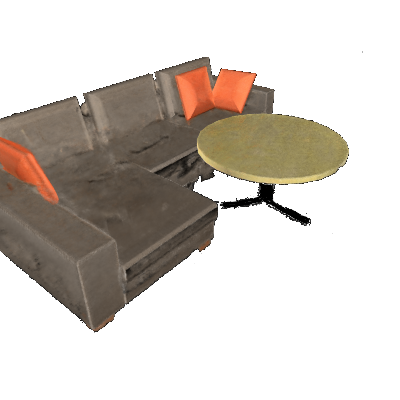}
  \caption*{Max. View Distance}
\end{subfigure}
\begin{subfigure}{0.45\columnwidth}
  \includegraphics[width=\columnwidth, height=2.7cm]{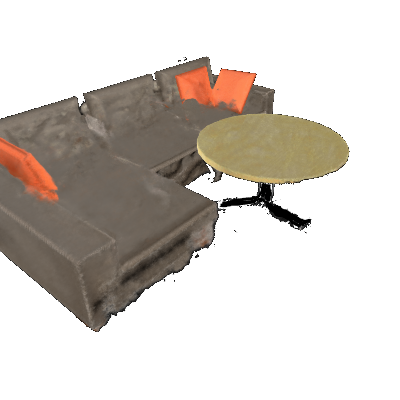}
  \caption*{Random}
\end{subfigure}
\begin{subfigure}{0.45\columnwidth}
  \includegraphics[width=\columnwidth, height=2.7cm]{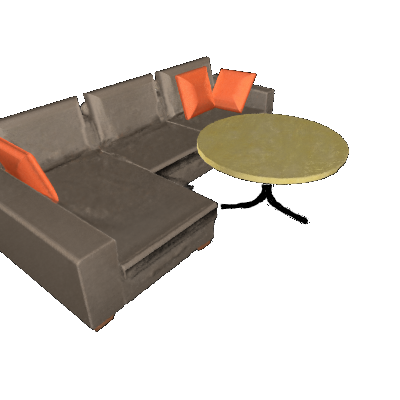}
  \caption*{Ours}
\end{subfigure}
\caption{Qualitative rendering results of trained NeRF models using image collections from three different NBV planners. We select a challenging view from each scene to show the rendering quality differences. Our uncertainty-guided planner adapts the acquisition of new measurements according to the scene structure in contrast to the heuristic strategies. For example, our approach takes more images of the car bonnet, 
which has more structural details compared to the car sides and is thus difficult to model using sparse views.}
\label{F: nerf rendering}
\vspace{-0.6cm}
\end{figure*}

After online NBV planning experiments in our simulator, described in~\cref{SS:NBV}, we use Instant-NGP~\citep{mueller2022} to train NeRF models using images collected by the three planning approaches, respectively, under the same training conditions. To evaluate the training results, we render $100$ test views using the trained NeRF models. We report the rendering metrics averaged over all experiment runs in~\cref{T:nerf_table} and show examples of rendering results at complex views from the scene in~\cref{F: nerf rendering}. Both quantitative and qualitative results verify that our planning strategy for collecting informative images boosts NeRF performance with limited training data. This indicates the benefits of using our approach to efficiently explore an unknown scene and collect informative images online. The 3D modelling of the scene can be done by training NeRFs offline, after a robotic mission, when computational resources are less constrained. 
\begin{table}[h]
\vspace{0.19cm}
\centering
\caption{NeRF training results using images collected from our planning experiments in the simulator. Best results in bold.}
\label{T:nerf_table}
\captionsetup{width=0.99\columnwidth}
\begin{tabularx}{0.99\columnwidth}{@{}cccc@{}}
\toprule
&& Car & Indoor\\ \midrule
 \multirow{3}{*}{PSNR $\uparrow$} 
  &Max. View Distance  & $27.37 \pm 0.65$ & $30.02 \pm 0.55$  \\
  &Random  & $25.73 \pm 0.83$ & $28.46 \pm 0.92$\\
  &Ours    & $\mathbf{28.35 \pm 0.53}$& $\mathbf{30.46 \pm 0.24}$\\ \midrule
 \multirow{3}{*}{SSIM $\uparrow$} 
  & Max. View Distance & $0.925 \pm 0.004 $& $0.937 \pm 0.003$  \\
  &Random    & $0.908 \pm 0.012$& $0.920 \pm 0.007$\\
  &Ours       & $\mathbf{0.934 \pm 0.004}$& $\mathbf{0.941 \pm 0.003}$ \\
 \bottomrule
\end{tabularx}
\end{table}
\section{Conclusions and Future Work} \label{S:conclusions}
In this work, we propose a novel mapless NBV planning framework for online robotic applications. We integrate uncertainty estimation in image-based neural rendering and exploit the predicted uncertainty to guide our measurement acquisition. We show that our uncertainty estimation is informative to the rendering quality at novel views and generalises to new scenes. Our planning experiments prove that our uncertainty-guided NBV planning scheme effectively finds informative views in an unknown scene. 
Image collection using our approach leads to more accurate scene representations via online image-based neural rendering and offline implicit reconstruction using NeRFs.
 
One limitation of our current framework is that rendering a high-resolution per-pixel uncertainty or RGB is inefficient for applications that require fast robot motion. To address this, future work will consider exploiting depth measurements to achieve more efficient sampling, thus speeding up the inference of neural rendering. To extend our framework to complex and cluttered environments, we plan to incorporate geometric uncertainty estimation for planning in unconstrained action spaces. Finally, we will investigate integrating semantic prediction with uncertainty estimation to enable exploring regions of interest in an unknown scene in applications where targeted inspection is necessary.

\section*{Acknowledgement}
We would like to thank Matteo Sodano for proofreading.

\bibliographystyle{IEEEtranN}
\footnotesize
\bibliography{2023-iros-jin}

\end{document}